\documentclass[conference]{IEEEtran}
\IEEEoverridecommandlockouts
\usepackage{cite}
\usepackage{amsmath,amssymb,amsfonts}
\usepackage{algorithmic}
\usepackage{graphicx}
\usepackage{textcomp}
\usepackage{xcolor}
\def\BibTeX{{\rm B\kern-.05em{\sc i\kern-.025em b}\kern-.08em
    T\kern-.1667em\lower.7ex\hbox{E}\kern-.125emX}}

\usepackage{algorithm}
\usepackage{amssymb}
\usepackage{amsmath}
\usepackage{multirow}
\usepackage{booktabs}
\usepackage{hyperref}

\begin{document}

\title{STP4D: Spatio-Temporal-Prompt Consistent Modeling for Text-to-4D Gaussian Splatting}

\author{
  \IEEEauthorblockN{Yunze Deng$^{1}$, Haijun Xiong$^{1}$, Bin Feng$^{^\dagger1}$, Xinggang Wang$^{1}$, Wenyu Liu$^{1}$}
  \IEEEauthorblockA{$^1$School of Electronic Information and Communications, Huazhong University of Science and Technology, Wuhan, China \\
  \{yunzedeng, xionghj, fengbin, xgwang, liuwy\}@hust.edu.cn}
  \thanks{$^\dagger$ To whom the correspondence should be addressed. This work was supported by the National Natural Science Foundation of China (No. 62376102).}
  }

\maketitle

\begin{abstract}
Text-to-4D generation is rapidly developing and widely applied in various scenarios. However, existing methods often fail to incorporate adequate spatio-temporal modeling and prompt alignment within a unified framework, resulting in temporal inconsistencies, geometric distortions, or low-quality 4D content that deviates from the provided texts. Therefore, we propose STP4D, a novel approach that aims to integrate comprehensive spatio-temporal-prompt consistency modeling for high-quality text-to-4D generation. Specifically, STP4D employs three carefully designed modules: Time-varying Prompt Embedding, Geometric Information Enhancement, and Temporal Extension Deformation, which collaborate to accomplish this goal. Furthermore, STP4D is among the first methods to exploit the Diffusion model to generate 4D Gaussians, combining the fine-grained modeling capabilities and the real-time rendering process of 4DGS with the rapid inference speed of the Diffusion model. Extensive experiments demonstrate that STP4D excels in generating high-fidelity 4D content with exceptional efficiency (approximately 4.6s per asset), surpassing existing methods in both quality and speed.
\end{abstract}


\section{Introduction}
\label{sec:intro}

Text-to-4D generation, a technology for creating interactive dynamic content, is promising in applications such as game development, film industry, virtual avatars, and VR/AR. However, it still faces significant challenges, including spatio-temporal inconsistency, textual misalignment, and non-real-time generation, which limit its broader application~\cite{liu2024comprehensive}.

\begin{figure*}[htp]
    \centering
    \includegraphics[width=1\linewidth]{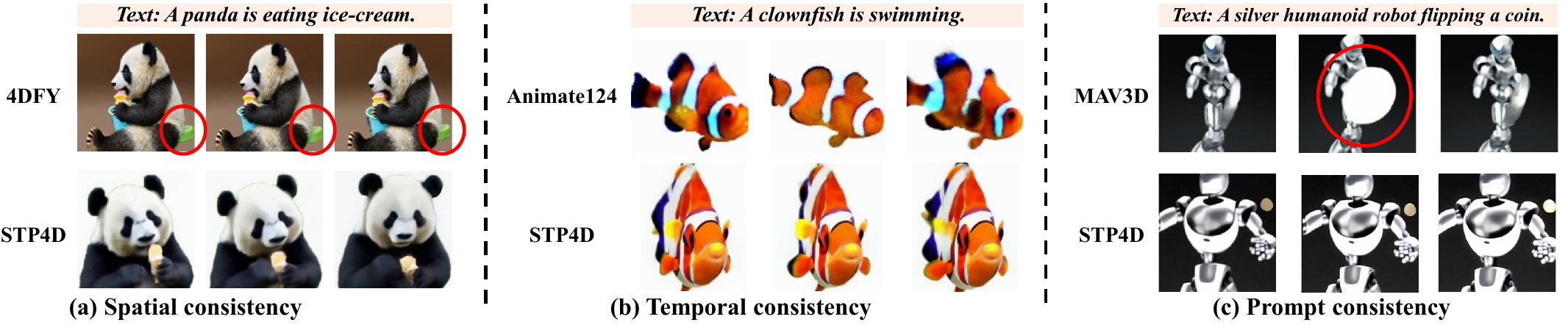}
    \caption{Visual comparisons between STP4D and other methods highlight the superiority of spatio-temporal-prompt consistency. (a) The panda generated from 4DFY, marked by the red circle, exhibits unreasonable body structures; (b) The clownfish from Animate124 shows drastic appearance changes across frames during swimming; (c) The robot generated by MAV3D misinterprets the prompt, failing to correctly throw and flip the coin.}
    \label{fig: intro}
\end{figure*}

Generating high-quality 4D content, both objectively and subjectively, is the primary goal of text-to-4D generation. However, temporal inconsistency may result in incoherent, unsmooth, or shaky content~\cite{yang2024deformable, huang2024sc}, while spatial inconsistency can lead to variations in geometric properties across different timestamps~\cite{liang2024diffusion4d}. Both of them may reduce the objective quality of the generation. On the other hand, prompt inconsistency can negatively impact subjective quality, causing misalignment between prompts and the generated content~\cite{liao2024clip}. Therefore, comprehensive spatio-temporal-prompt modeling is crucial for high-quality and high-fidelity generation.

Existing solutions often fail to achieve adequate spatial-temporal-prompt modeling simultaneously, leading to inconsistent generation. Some typical methods tend to ignore the modeling of prompts, with certain approaches highly relying on guidance from text-to-image/video (T2I/T2V) models for alignment~\cite{yi2023gaussiandreamer, bahmani20244d, wang2024prolificdreamer}. This underestimation of prompt modeling may hinder the generation of semantically correlated content, while some methods even fail when the guiding models are poor. Moreover, other methods may lack comprehensive spatio-temporal modeling, leading to low-quality generation. For example,~\cite{singer2023text} only utilizes HexPlane~\cite{cao2023hexplane} for spatio-temporal modeling, while SC-GS \cite{huang2024sc} relies solely on MLP to embed temporal information. To address them, we propose a unified framework that integrates spatio-temporal-prompt modeling, improving both the objective and subjective quality of generated 4D content, as shown in Fig.\ref{fig: intro}.

Based on the analysis above, this paper introduces a novel text-to-\textbf{4D} method that ensures \textbf{S}patio-\textbf{T}emporal-\textbf{P}rompt consistency, referred to as \textbf{STP4D}. To achieve comprehensive consistency modeling, STP4D incorporates three carefully designed modules: Time-varying Prompt Embedding (TPE), Geometric Information Enhancement (GIE), and Temporal Extension Deformation (TED), respectively. Specifically, the TPE module embeds more accurate prompt information into Gaussians for each frame, ensuring high alignment with the prompts. The GIE module employs the specially designed GroupFormer to extract inter-group and intra-group spatio-temporal features, enhancing the geometric information of the Gaussians. The TED module extends content from anchor frames to actual frames via cross-attention. These three modules work collaboratively, incorporating multiple consistency constraints to achieve robust and comprehensive spatio-temporal-prompt modeling.

Furthermore, to enhance framework efficiency, our method integrates the Diffusion model with 4D Gaussian Splatting (4DGS). Previous studies demonstrate that Denoising Diffusion Implicit Models (DDIM) possess robust generative capabilities, efficient conditional embedding, and rapid inference~\cite{croitoru2023diffusion, cao2024survey, song2020denoising}. Meanwhile, 4DGS~\cite{wu20244d, duan20244d} enables fine-grained 4D content representation and extremely fast rendering. Therefore, our method exploits DDIM to generate 4DGS with prompt information as conditions, harnessing the strengths of both DDIM and 4DGS. This allows STP4D to generate high-quality 4D assets with an impressive inference speed of only 4.6s per asset (with 24 frames). Notably, STP4D is among the first methods to generate 4DGS using the Diffusion model. In summary, our contributions are as follows:
\begin{itemize}
    \item We propose a novel text-to-4D network, STP4D, which performs comprehensive spatio-temporal-prompt modeling through the TPE, GIE, and TED modules collectively. Moreover, carefully designed spatio-temporal-prompt constraints further enhance the consistency.
    \item STP4D is among the first methods to explore the use of Diffusion models for directly generating high-quality dynamic Gaussians, enabling rapid inference, fine-grained representations, and fast rendering.
    \item We conduct quantitative experiments on the Diffusion4D dataset and qualitative visual comparisons with other methods. The results highlight the state-of-the-art performance and rapid generation capability ($\sim$4.6s per asset) of STP4D, while ablation studies validate the effectiveness of each module.
\end{itemize}

\begin{figure*}[htp]
    \centering
    \includegraphics[width=1\linewidth]{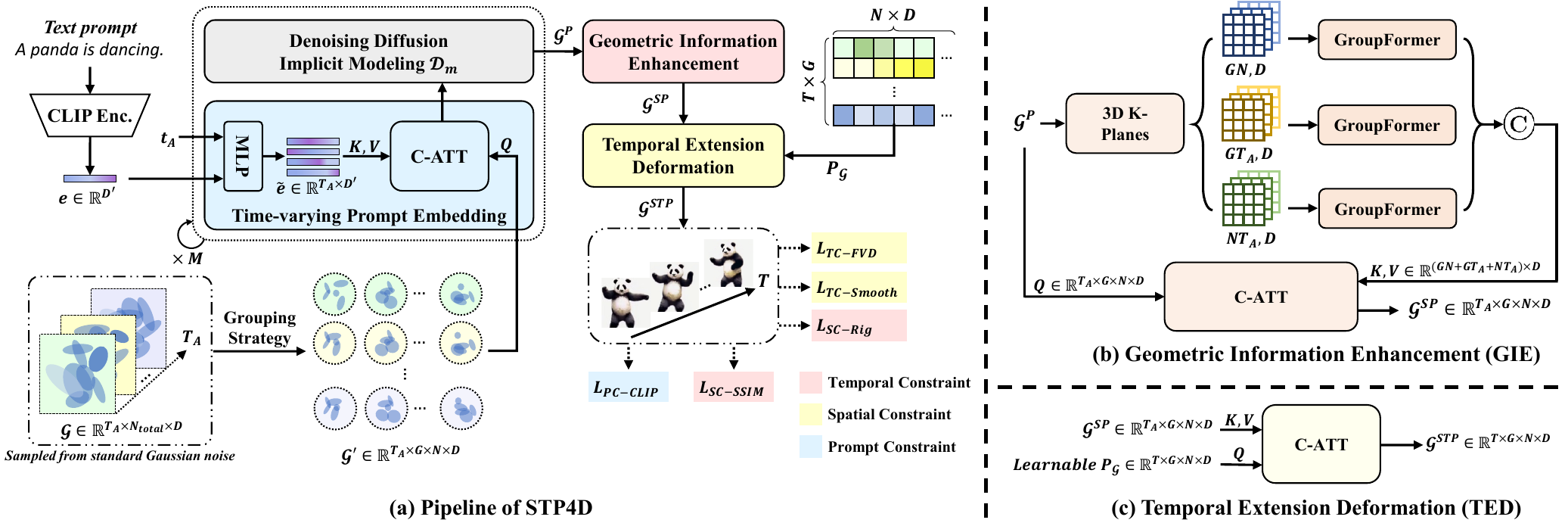}
    \caption{(a) Pipeline of STP4D; (b) Details of Geometric Information Enhancement (GIE); (c) Details of Temporal Extension Deformation (TED).
    }
    \label{fig: overview}
\end{figure*}

\section{Related Work}

\textbf{Text-to-4D.} Currently, most text-to-4D methods rely heavily on T2I/T2V models for guidance to optimize 4D representations, \textit{e.g.} Neural Radiance Field (NeRF)~\cite{mildenhall2021nerf}, mesh or 4DGS.
Techniques such as Score Distillation Sampling (SDS)~\cite{huang2020probability, poole2022dreamfusion} are commonly used to distill knowledge from guidance models for effective 4D generation. 
MAV3D~\cite{singer2023text}, the first text-to-4D method, utilizes SDS derived from video Diffusion models to modify 4D NeRF.
Animate124~\cite{zhao2023animate124} optimizes 4D grid dynamic NeRF in stages with multiple Diffusion priors.
AYG~\cite{ling2024align}, the first text-to-4D method based on Gaussian Splatting, employs SDS from T2I and T2V Diffusion models to optimize 4DGS from static 3D scenes.
However, these methods often overlook spatio-temporal-prompt modeling capabilities of themselves and require substantial time to optimize NeRF or 4DGS for generation. STP4D exploits the Diffusion model to generate 4DGS directly, without the necessity for a prior guidance model, while enabling rapid inference.

\textbf{Spatio-Temporal-Prompt Modeling.} Existing 3D and 4D generation methods have achieved significant success in spatial consistency, temporal consistency, and prompt embedding. However, there is no method that has fully considered comprehensive spatio-temporal-prompt consistency modeling simultaneously. Dream-in-4D~\cite{zheng2024unified} uses video Diffusion guidance for motion modeling, but lacks geometric modeling and constraints, failing to ensure spatial consistency. SC-GS~\cite{huang2024sc} only uses MLPs to predict deformation fields of static Gaussians to generate 4D Gaussians, without sufficient temporal modeling. STAG4D~\cite{zeng2024stag4d} employs Deformation Planes and a multi-view Diffusion model for multi-view spatio-temporal consistency, but its strategy of prompt modeling is limited, neglecting prompt alignment. In contrast, STP4D achieves comprehensive spatio-temporal-prompt modeling, ensuring high quality in both objective and subjective aspects of the generated content.

\section{Method}

\subsection{Pipeline}

Fig.\ref{fig: overview} illustrates the proposed end-to-end text-to-4D framework, STP4D, where three primary modules (\textit{i.e.} TPE, GIE, and TED) collaborate spatio-temporal-prompt modeling. During both training and inference, the attributes of the initialized Gaussians are sampled from standard Gaussian noise.
STP4D takes text prompts $tx$ as input, initializing $N_{total}$ Gaussians $\mathcal{G} \in \mathbb{R}^{T_A \times N_{total} \times D}$ with $T_A$ anchor frames and $D$ optimizable attributes. 
First, the Gaussians are grouped using a KNN strategy, clustering spatially close Gaussians into $G$ groups, resulting in $\mathcal{G'} \in \mathbb{R}^{T_A \times G \times N \times D}$. 
CLIP~\cite{radford2021learning} model $E_{clip}$ is taken as a text encoder to obtain prompt information $ e=E_{clip}(tx) \in \mathbb{R}^{D'}$, where $D'$ is the dimension of latent features. Then, an $M$-step DDIM $\mathcal{D}_m$ iteratively denoises $\mathcal{G'}$, embedding prompt information $e$ at each step $m$:
\begin{equation}      
\label{eq: ddim}
\mathcal{G}_e(m)=\mathcal{D}_m(\mathrm{TPE}(\mathcal{G}_e(m-1), e))\in \mathbb{R} ^{T_A\times G\times N\times D},
\end{equation}
where $m = 1, ..., M$, with $\mathcal{G'}$ as $\mathcal{G}_e(0)$. $\mathcal{G}_e(m)$ represents the predicted Gaussians at each step $m$.
After denoising, the predicted Gaussians $\mathcal{G}^P = \mathcal{G}_e(M)$ are obtained, highly aligned with the prompts. The GIE module then utilizes K-Planes~\cite{fridovich2023k} and GroupFormer to model inter-group and intra-group spatio-temporal features, enhancing geometric information, yielding $\mathcal{G}^{SP}$.
Finally, the TED module extends the anchor frames to the actual $T$ frames using a learnable weight pool, generating the 4D Gaussians $\mathcal{G}^{STP}$. Multiple loss functions ensure the spatio-temporal-prompt consistency of $\mathcal{G}^{STP}$.

\subsection{Time-varying Prompt Embedding (TPE)}

The TPE module is designed to ensure that the generated 4D content highly aligns with the prompt information, as illustrated in Fig.\ref{fig: overview} (a). For text-to-4D generation, more targeted text features for each frame can lead to more accurate and anticipated results. 
We map the text features $e \in \mathbb{R}^{D'}$ to $T_A$ frames, forming the time-varying prompt information $\tilde{e}\in \mathbb{R}^{T_A \times D'}$, which is embedded into the corresponding frame of Gaussians. Specifically, during the $m$-th denoising step, the process involving the time-varying prompt information $\tilde{e}$ and the TPE module can be expressed as:

\begin{equation}
\begin{cases}
\tilde{e}=\mathrm{MLP}(e,t_A) \in \mathbb{R}^{T_A \times D'}, t_A=1,\cdots ,T_A \\
\mathrm{TPE}(\mathcal{G}_e(m-1),e)=\mathrm{C \text{-} ATT}(\mathcal{G}_e(m-1),\tilde{e},\tilde{e})
\end{cases}.
\end{equation}
Here, $\mathrm{C \text{-} ATT}(Q, K, V)$  represents the cross-attention mechanism, where $Q$, $K$ and $V$ denote Query, Key and Value respectively. 
According to \eqref{eq: ddim}, DDIM utilizes TPE to embed time-varying prompt information $\tilde{e}$ into the Gaussians during each denoising step, ultimately obtaining the predicted Gaussians that are highly aligned with the text description.

\subsection{Geometric Information Enhancement (GIE)}

The GIE module models both global and local spatio-temporal features of Gaussians to enhance geometric information and improve representation, as shown in Fig.\ref{fig: overview}(b). To reduce computational cost and enable global modeling, we represent the Gaussians $\mathcal{G}^{P}$ as 3D voxels and decompose them into three low-rank matrices using the K-Planes strategy~\cite{fridovich2023k}, denoted as
$\mathcal{G}^{P}_{1}\in \mathbb{R}^{GN\times D}$,
$\mathcal{G}^{P}_{2}\in \mathbb{R}^{G T_A\times D}$,
and $\mathcal{G}^{P}_{3}\in \mathbb{R}^{N T_A\times D}$.
Inspired by~\cite{duan2024condaformer}, we design GroupFormer to extract Gaussian features, denoted as $\mathrm{GF}_i(i\in \{1,2,3\})$. 
Implementation details of both K-Planes and GroupFormer are provided in the \textbf{supplementary materials}.
The spatio-temporal features from the three planes are then fused as:

\begin{equation}
    \mathcal{G}^{SP}_f=\mathrm{Cat}(\mathrm{GF}_1(\mathcal{G}^{P}_{1}), \mathrm{GF}_2(\mathcal{G}^{P}_{2}), \mathrm{GF}_3(\mathcal{G}^{P}_{3})),
\end{equation}
where $\mathrm{Cat}(\cdot)$ denotes concatenation. GroupFormer enables efficient global and local spatio-temporal modeling of both inter-group and intra-group features of Gaussians. Finally, these extracted features are used to geometrically enhance the Gaussians $\mathcal{G}^{P}$ via cross-attention:
\begin{equation}
    \mathcal{G}^{SP}=\mathrm{C \text{-} ATT}(\mathcal{G}^{P}, \mathcal{G}^{SP}_f, \mathcal{G}^{SP}_f) \in \mathbb{R}^{T_A\times G\times N\times D}.
\end{equation}

\subsection{Temporal Extension Deformation (TED)}

\begin{table*}[htbp]
    \centering
    \caption{Quantitative results of various text-to-4D methods on Diffusion4D test set and human performance, where the bold values represent the best results. $\mathcal{T}$ and $\mathcal{I}$ represent the training time, and inference time for each asset, respectively.}
    \begin{tabular}{ccccc|ccccc}
\toprule
\multicolumn{1}{c}{}  & \multicolumn{4}{c|}{Metrics}                     & \multicolumn{5}{c}{Human Performance} \\
Method       & CLIP-F↑ & CLIP-O↑ & FVD↓ & $(\mathcal{T}, \mathcal{{I}})$↓ & 3DC   & AQ   & MF   & TA   & OA  \\ \midrule
MAV3D~\cite{singer2022make}  &  0.745  &  0.556  & 549.8    & (-, $\sim$6.5h)        &    5\%   &   4\%   &    5\%  &   9\%   &     4\%     \\
Animate124~\cite{zhao2023animate124}   & 0.752   & 0.565   & 584.2    & (-, $\sim$9h)        &   10\%    &   7\%   &    15\%  &    11\%  &     10\%     \\
4DFY~\cite{bahmani20244d}         & 0.792   & 0.622   & 531.0    & (-, $\sim$23h)       &     14\%  &   15\%   &   11\%   &  14\%    &      13\%    \\
Diffusion4D~\cite{liang2024diffusion4d}  & 0.810   & 0.650   & 482.6    & (-, $\sim$8m)        &    26\%   &    29\%  &   22\%   &    20\%  &     24\%     \\
\textbf{STP4D (Ours)}  & \textbf{0.841}   & \textbf{0.705}   & \textbf{441.1}    & ($\sim$11h, \textbf{$\sim$4.6s})      &    \textbf{45\%}   &   \textbf{45\%}   &   \textbf{47\%}   &   \textbf{46\%}   &      \textbf{49\%}    \\ \bottomrule
\end{tabular}
    \label{tab: sota}
\end{table*}

As shown in Fig.\ref{fig: overview}(c), the TED module transitions from anchor frames to actual frames by globally modeling spatio-temporal features across Gaussian groups for all anchor frames, avoiding the computational cost of modeling Gaussians for all actual frames while maintaining high-fidelity multi-frame 4D generation. The module treats the $T_A$ anchor frames in $\mathcal{G}^{SP}$ as bases and extends them to $T$ frames with a ratio $\eta = T/T_A$, using a learnable weight pool $P_{\mathcal{G}} \in \mathbb{R}^{T \times G \times N \times D}$ to provide weights. Final Gaussians with $T$ frames are generated by cross-attention:

\begin{equation}
    \mathcal{G}^{STP}=\mathrm{C \text{-} ATT}(P_{\mathcal{G}}, \mathcal{G}^{SP}, \mathcal{G}^{SP}) \in \mathbb{R}^{T\times G\times N\times D}.
\end{equation}

\subsection{Spatial-Temporal-Prompt Constraints}

STP4D employs multiple constraints to ensure spatio-temporal-prompt consistency in the generated content. Here, we briefly introduce the function of different constraints and their fusion strategy. For detailed design specifications, please refer to the \textbf{supplementary materials}.

\textbf{Spatial constraints.} $L_{SC-SSIM}$ minimizes spatial structural differences between the rendered image and ground truth using SSIM \cite{wang2004image}, enhancing the geometric realism of the generated content. Inspired by \cite{sorkine2007rigid}, we design $L_{SC-Rig}$ to enforce local rigidity within each Gaussian group, preventing unreasonable deformations and ensuring spatial consistency.

\textbf{Temporal constraints.} $L_{TC-FVD}$ minimizes the FVD \cite{heusel2017gans} score of the rendered sequences, promoting smoother, higher-quality generation. $L_{TC-Smooth}$ uses the Savitzky-Golay \cite{dombi2020adaptive} filter to measure the smoothness of each Gaussian’s attributes across frames, reducing the risk of incoherent content.

\textbf{Prompt constraints.} $L_{PC-CLIP}$ aligns the generated 4D content with the provided text description by calculating the CLIP score between the rendered image of each frame and the corresponding text.

\textbf{Fusion Strategy.} We adopt the following formula to implement the loss function fusion:
\begin{equation}
\begin{aligned}
L=& \theta_1 L_{SC-SSIM}+\theta_2 L_{SC-Rig}+\theta_3 L_{TC-FVD}+  
\\ & \theta_4 L_{TC-Smooth}+\theta_5 L_{PC-CLIP}.
\end{aligned}
\end{equation}
Based on our experience, we set the weight parameters for the fusion as $[\theta_1, \theta_2, \theta_3, \theta_4, \theta_5]=[1, 0.01, 0.001, 0.1, 1]$.

\section{Experiments}

In this section, we first introduce the evaluation metrics used in STP4D, followed by the analysis of experimental results and ablation studies. Network implementation details, dataset information, additional experiments, and further visualizations can be found in the \textbf{supplementary materials}. For a fair comparison, we follow~\cite{liang2024diffusion4d} by retaining 20 cases as a test set and using the remaining for training.

\subsection{Metrics}

For quantitative analysis, we evaluate our method on the Diffusion4D~\cite{liang2024diffusion4d} test set.
Following~\cite{liang2024diffusion4d}, we use CLIP-O and CLIP-F to measure the semantic correlation between generated content and text prompts.
Temporal consistency is assessed using the FVD~\cite{heusel2017gans} score to evaluate the quality and smoothness of the rendered sequences. Additionally, we compare training time $\mathcal{T}$ and inference time per asset $\mathcal{I}$ across methods.
For qualitative analysis, we present a series of visualizations for comparative analysis and conduct a user study with 40 participants following~\cite{liang2024diffusion4d}.
Participants independently evaluate anonymized video sequences generated by different models, rating five properties: 3D geometry consistency (3DC), appearance quality (AQ), motion fidelity (MF), text alignment (TA), and overall performance (OA), with results reported as human performance statistics.

\subsection{Experimental Results}

\textbf{Quantitative Results.} As shown in Tab.\ref{tab: sota}, our method outperforms the current state-of-the-art methods in all quantitative metrics. Specifically, STP4D surpasses the second-best method, Diffusion4D~\cite{liang2024diffusion4d}, by 0.031 and 0.055 in CLIP-F and CLIP-O scores, respectively, demonstrating its strong semantic alignment between prompts and 4D content. STP4D achieves 441.1 in FVD score, superior to all other methods, indicating that it can generate coherent, high-quality 4D assets. Notably, STP4D requires only 4.6s to generate a 4D asset during inference, making it around 
$100\times$ faster than Diffusion4D, suitable for some practical scenes that require fast 4D generation. These results highlight STP4D’s ability to efficiently produce high-quality, spatio-temporal-prompt consistent 4D content, establishing its state-of-the-art performance.

\begin{table}[t]
    \centering
    \caption{Study of the effectiveness of three modules in STP4D in terms of the quantitative metrics on Diffusion4D.}
    \begin{tabular}{cccc}
    \toprule
Method        & CLIP-F↑ & CLIP-O↑ & FVD↓ \\ \midrule
\textbf{STP4D }   & \textbf{0.841}   & \textbf{0.705}   & \textbf{441.1}   

    \\
 \textit{w/o} TPE &     0.805    &     0.672    &     447.5     \\
 \textit{w/o} GIE &   0.792      &     0.655    &     496.7     \\
 \textit{w/o} TED &    0.829     &     0.694    &    458.5      \\
 \bottomrule
    \end{tabular}
    \label{tab: ablation1}
\end{table}

\begin{figure*}[t]
    \centering
    \includegraphics[width=1\linewidth]{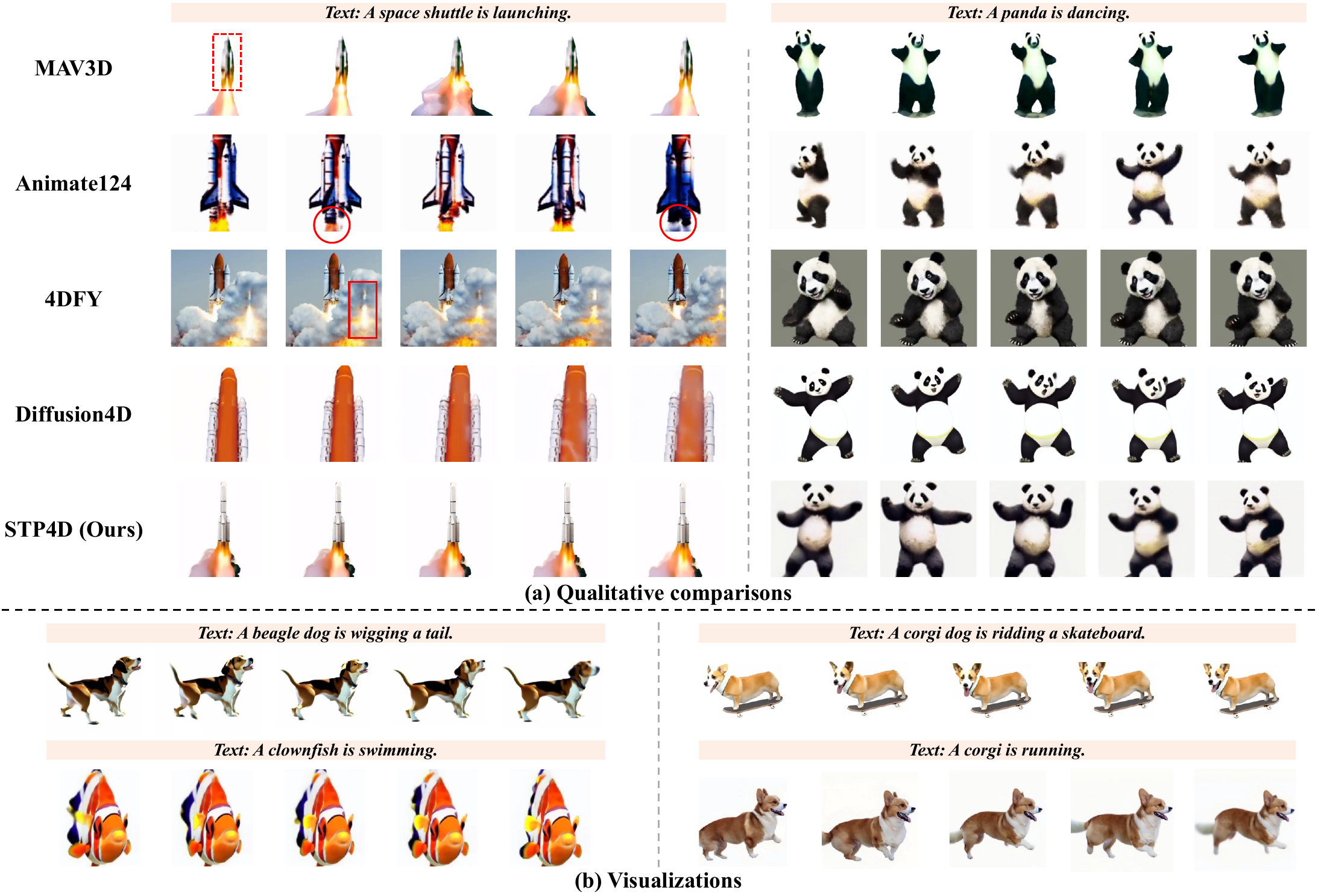}
    \caption{(a) Visual comparisons between STP4D and other competitive methods. (b) Various 4D assets generated from STP4D.
    }
    \label{fig: qualitative}
\end{figure*}

\textbf{Qualitative Results.} Furthermore, we compare the visualizations of STP4D with other methods in Fig.\ref{fig: qualitative}(a). Qualitative comparisons show that STP4D successfully generates high-quality 4D content that strongly aligns with prompts, demonstrating its superior performance.
Taking \textit{``A space shuttle is launching''} as an example, as presented in the red-dashed box of Fig.\ref{fig: qualitative}(a), the space shuttle generated by MAV3D~\cite{singer2022make} exhibits a blurred appearance and lacks geometric structure. 
The red-circled area highlights the unsmooth content generated by Animate124~\cite{zhao2023animate124} across different frames. 
Moreover, 4DFY~\cite{bahmani20244d} deviates from texts and produces more than one space shuttle, as shown in the red-solid box, indicating that 4DFY does not comprehensively interpret the prompt. In contrast, the space shuttle generated by STP4D features clear structure, smooth motion, and accurate alignment with texts. Fig.\ref{fig: qualitative}(b) displays several assets generated by STP4D, with more visualizations and GIFs available in the \textbf{supplementary materials}.

\begin{table}[t]
    \centering
    \caption{
    Study of the effectiveness of three proposed constraints in terms of the quantitative metrics on Diffusion4D.
     }
    \begin{tabular}{cccc}
    \toprule
Method                    & CLIP-F↑ & CLIP-O↑ & FVD↓ \\ \midrule
\textbf{STP4D}               & \textbf{0.841}   & 0.705   & \textbf{441.1}   
\\
 \textit{w/o} $L_{SC-Rig}$    &     0.802    &    0.661     &     473.0     \\
 \textit{w/o} $L_{TC-FVD}$    &    0.825     &     0.684    &     498.0     \\
 \textit{w/o} $L_{TC-Smooth}$ &     0.838    &    \textbf{0.709}     &     485.7     \\
 \bottomrule
\end{tabular}
    \label{tab: ablation2}
\end{table}

\textbf{User Study.} We conduct a user study involving 40 participants and find that people exhibit a stronger preference for our method across different properties. As presented in Tab.\ref{tab: sota}, more participants favor STP4D in terms of 3DC, AQ, and MF, demonstrating its ability to generate 4D assets with realistic appearance, detailed geometry, and smooth motion. High TA scores further confirm STP4D's excellent alignment with text prompts. Overall, STP4D can generate high-quality content both objectively and subjectively, earning greater approval from people.

\subsection{Ablation Study}

\textbf{Effectiveness of Main Components.} We conduct an effectiveness study on the three main modules of STP4D: TED, GIE, and TPE, with results presented in Tab.\ref{tab: ablation1}.
To fully validate their contribution, we remove TPE entirely and allow the Diffusion model to directly use the text features $e$ extracted by the CLIP encoder as conditions. For TED, we remove it and let the entire network directly predict the Gaussians with the length of actual frames $T$. 
Results show that removing TPE reduces the CLIP-F and CLIP-O scores by 0.036 and 0.033, respectively, indicating that TPE embeds more accurate prompt information, improving prompt consistency.
Removing GIE significantly worsens performance across all metrics by -0.049, -0.050, and +55.6, highlighting its importance in enhancing the geometric representation of 4DGS using global and local spatio-temporal features.
Removing TED leads to a slight performance drop, suggesting that TED successfully achieves both consistency modeling and the transition from anchor to actual frames. These modules collaboratively ensure the spatio-temporal-prompt consistency of STP4D.

\textbf{Effectiveness of Three Spatio-temporal Constraints.} We conduct an ablation study on the proposed losses, with results shown in Tab.\ref{tab: ablation2}. 
Notably, we did not conduct ablation on $L_{PC-CLIP}$ and $L_{SC-SSIM}$, as both are essential during training.
Removing $L_{SC-Rig}$ reduces performance on CLIP-F, CLIP-O, and FVD by -0.039, -0.044, and +31.9, respectively, highlighting the significance of local rigidity in preserving fine-grained geometric features for high-quality 4D assets. 
Additionally, despite the minor negative impact on the CLIP-O score, $L_{TC-Smooth}$ significantly enhances the overall network performance.
Removing $L_{TC-FVD}$ and $L_{TC-Smooth}$ increases FVD by +56.9 and +44.6, respectively, demonstrating their effectiveness in constraining temporal consistency.
 
In the \textbf{supplementary materials}, we provide experimental results on the temporal extension ratio $\eta$ and the exploration of the generality of $\eta=2:1$. Additionally, we investigate the impact of the Gaussian grouping strategy on network performance. More visualizations are also included.

\section{Conclusion}
We propose STP4D, a novel end-to-end text-to-4D method that integrates three carefully designed modules collectively to ensure spatio-temporal-prompt consistency in generated 4D assets. As a novel method to exploit Diffusion models for generating 4DGS, STP4D offers both rapid inference and high-quality generation simultaneously. Experimental results demonstrate its superiority over existing methods and validate the effectiveness of its modules and loss functions.

\section{Supplements of STP4D Method}
\label{sec:s1}

\subsection{Preliminary}

In this subsection, we briefly review the representation and rendering of Gaussian Splatting, as well as the DDIM.

\textbf{Generative Gaussian Splatting.} Anisotropic 3D Gaussians can explicitly represent a realistic 3D scene by optimizing each Gaussian's attributes.
Specifically, for a set of Gaussians $\mathcal{G}$, the attributes of the $i$-th Gaussian are $\mathcal{G}_i=\{x_i, r_i, s_i, c_i, \sigma_i\}$, where $x_i \in \mathbb{R}^3$ is the central position in 3D space, $c_i$ and $\sigma_i \in \mathbb{R}^1$ are color and opacity.
$r_i$ and $s_i$ represent a quaternion and a 3D vector, respectively, derived from the rotation matrix $R \in \mathbb{R}^{3\times 3}$ and the scaling matrix $S \in \mathbb{R}^{3\times 3}$.
These two matrices can form the covariance matrix $\Sigma_i = RSS^TR^T$.
Given a pixel $p$, the $i\mathrm{-th}$ 3D Gaussian primitive $\mathcal{G}_i$ can be described as:
\begin{equation}
    \mathcal{G}_i(p)=exp^{-\frac{1}{2} x_i \Sigma_i x_i^T}.
\end{equation}
Using rendering method from \cite{kopanas2022neural, kerbl20233d}, Gaussians $\mathcal{G}$ are projected into image space:
\begin{equation}
    \label{eq-render}
    C=\sum_{i}c_i \alpha_i\prod_{j=1}^{i-1}(1-\alpha_j),
\end{equation}
where $\alpha_i$ is rendering opacity computed by $\sigma_i$. The proposed STP4D utilizes 4DGS to represent dynamic 3D content and employs \eqref{eq-render} for rendering.

\textbf{Denoising Diffusion Implicit Models.} DDIM is built upon the Denoising Diffusion Probabilistic Model (DDPM) \cite{ho2020denoising, sohl2015deep} and generalizes it by providing a more efficient sampling process. Compared to DDPM, DDIM offers faster sampling speeds, and the sampling process is non-Markovian. The original data before noising is $x_0$, and the distribution of the data after $t$ steps of noising is $x_t$, where $t=1,...,T$. Let $q$ represent the distributions, and the denoising process in DDIM recovers $x_0$ from $x_t$ by:
\begin{equation}
\begin{aligned}
    & q(x_{t-1} \mid x_t, x_0)=\mathcal{N}(x_{t-1};\mu,\tilde{ \beta}_t\mathrm{I}), where
    \\ & \mu=\sqrt{\bar{\alpha}_{t-1}}x_0+\sqrt{1-\bar{\alpha}_{t-1}-\tilde{\beta}_t} \cdot \frac{x_t-\sqrt{\bar{\alpha}_{t}}x_0}{\sqrt{1-\bar{\alpha}_t}}. 
\end{aligned}
\end{equation}
Here, $\alpha_t$ is the noising parameter, and $\bar{\alpha_t}=\prod_{i=1}^{t}\alpha_i$. In DDIM, $\tilde{\beta}_t$ is set to 0, making the reverse Diffusion process deterministic. Consequently, given the random noise $x_T$, the process yields a unique sampling result $x_0$. In our paper, the denoising process at each step of an $M$-step DDIM is simplified as $\mathcal{D}_m(\cdot)$, where $m=1,...,M$ is the current denoising step.

\subsection{Details of 3D K-Planes and GroupFormer}

Due to the high computational cost of directly modeling 4D Gaussians, we introduce K-Planes \cite{fridovich2023k} technique to decompose them into three low-rank matrices for efficient spatio-temporal modeling. Specifically, we treat three dimensions $(T_A, G, N)$ of the 4D Gaussians $\mathcal{G}^{P} \in \mathbb{R}^{T_A \times G \times N \times D}$ as a 3D space. K-Planes allows this 3D space to be represented by three 2D planes: $(GN, GT_A, NT_A)$. With this decomposition strategy, the 4D Gaussians are split into three low-rank matrices: $\mathcal{G}^{P}_{1} \in \mathbb{R}^{GN \times D}$, $\mathcal{G}^{P}_{2} \in \mathbb{R}^{G T_A \times D}$, and $\mathcal{G}^{P}_{3} \in \mathbb{R}^{N T_A \times D}$. These three low-rank matrices are significant for GroupFormer to efficiently perform spatio-temporal modeling.

\begin{figure}[t]
    \centering
    \includegraphics[width=1\linewidth]{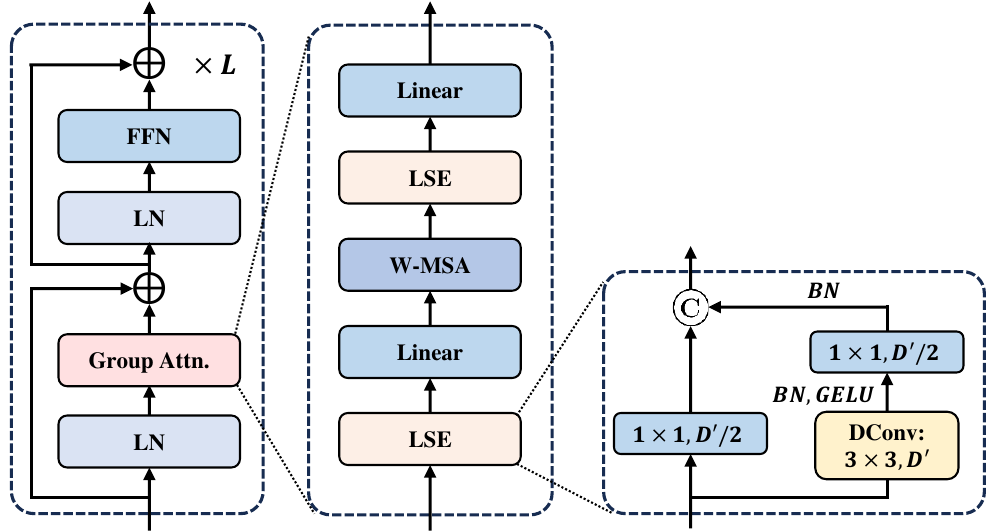}
    \caption{The detailed structure of GroupFormer. $D'$ represents the hidden dimension.
    }
    \label{fig: groupformer}
\end{figure}

GroupFormer enables global and local spatio-temporal modeling of both inter-group and intra-group with low computational cost, as shown in Fig.\ref{fig: groupformer}. The core of GroupFormer is GroupAttention, which is implemented by Window-based Multi-Head Self-Attention (W-MSA) \cite{liu2021swin} and Local Structure Enhancement (LSE) \cite{duan2024condaformer}. The former performs low-complexity self-attention within windows to extract local features, while the latter facilitates communication between windows using depth-wise sparse convolution of size $3 \times 3$, thereby achieving global feature extraction. Denoting W-MSA and LSE as $\mathrm{W \text{-} MSA(\cdot)}$ and $\mathrm{LSE(\cdot)}$ respectively, GroupAttention can be expressed as:

\begin{equation}
    \mathrm{GA}(\cdot)=\mathrm{Li}(\mathrm{LSE}(\mathrm{W \text{-} MSA}(\mathrm{Li}(\mathrm{LSE}(\cdot))))),
\end{equation}
where $\mathrm{Li}(\cdot)$ denotes linear layer. Each $\mathrm{GF}_i (i \in {1,2,3})$ consists of $L$ layers of GroupAttention, and the $j$-th layer can be represented as (taking $x_{j-1}$ as inputs, $j=1,...,L$):
\begin{equation}
\begin{aligned}
    & x'_{j} = \mathrm{GA}(\mathrm{LN}(x_{j-1}))+x_{j-1}, and\\ 
    &  \mathrm{GF}_i^j(x_{j-1}) = \mathrm{FFN}(\mathrm{LN}(x'_{j})) + x'_{j}. 
\end{aligned}
\end{equation}
Here $\mathrm{LN}(\cdot)$ and $\mathrm{FFN}(\cdot)$ denote layer-norm operation and feed forward network, respectively. 

\subsection{Loss functions}

We have provided a brief introduction of each constraint in our paper, thus this subsection will focus on their detailed design specifications. Notably, the value of each loss is minimized for optimal performance.

$\mathbf{L_{SC-SSIM}}$. We utilize SSIM \cite{wang2004image} to design a loss function that minimizes the structural difference between each frame of the rendered sequence and the ground truth sequence, ensuring generation with realistic geometric structures. The ground truth images rendered from dynamic 3D assets are used as references, denoted as $V_{gt}$, while the rendered images from STP4D (using the same camera pose) are taken as targets, represented as $V$. Each frame of $V_{gt}$ and $V$ can be written as $V^t_{gt}$ and $V^t$, where $t=1,…,T$, and $L_{SC-SSIM}$ can be obtained by:

\begin{equation}
L_{SC-SSIM}=\frac{1}{T} \sum_{t=1}^T{(1-\mathrm{SSIM}(V_{gt}^t, V^t))}.
\end{equation}

$\mathbf{L_{SC-Rig}}$. Inspired by \cite{sorkine2007rigid}, we design a spatial constraint $L_{SC-Rig}$ to enforce local rigidity within each group of Gaussians. The local rigidity of intra-group Gaussians can prevent unreasonable deformations and geometric distortions, enhancing spatial consistency. Specifically, for a set of $T$-frame Gaussians $\mathcal{G} \in \mathbb{R}^{T\times G\times N\times D}$, we randomly sample $K$ timestamp pairs. Taking the $k$-th pair $(t_1^k, t_2^k)(k=1,...,K)$ as an example, a Gaussian from each group is randomly selected as a reference point, with its position at time $t_1^k$ denoted as $x_i^{g, t_1^k }(g=1,…,G)$. At the same time, other points within the same group are denoted as $x_j^{g, t_1^k }(j=1,…,N, i \neq j)$. The local rigid loss can be expressed as:
\begin{equation}
\begin{aligned}
 L_{SC-Rig}= & \sum_{i=1}^{N}
 \sum_{k=1}^{K} \sum_{g=1}^{G} \sum_{j=1, j \neq i}^{N} 
 \\
 & \label{eqn3} ( \left \| x_i^{g, t_1^k }-x_j^{g, t_1^k }  \right \|_2 - \left \| x_i^{g, t_2^k }-x_j^{g, t_2^k }  \right \|_2   )^2.
\end{aligned}
\end{equation}

$\mathbf{L_{TC-FVD}}$. FVD score \cite{heusel2017gans} is employed to design a loss function that ensures smooth and high-quality generated content. We use a pre-trained I3D network to extract features from $V_{gt}$ and $V$, then apply the Frechet distance function $\mathcal{F}$ to calculate the spatial distribution difference between the generated video features and the real video features. This process is represented as :
\begin{equation}
L_{TC-FVD}=\mathcal{F}(I3D(V), I3D(V_{gt})).
\end{equation}

$\mathbf{L_{TC-Smooth}}$. $L_{TC-Smooth}$ employs filters to measure the smoothness of each Gaussian's attributes across all frames, reducing the risk of incoherent content. Specifically, the variation of a positional attribute of the same Gaussian across 
$T$ frames is denoted as the signal $g_{attr}^{i}$, where $i=1,...,N_{total}$. Next, a Savitzky-Golay \cite{dombi2020adaptive} filter $\mathcal{S}$ with a window size of 7 and a polynomial order of 3 is applied for $g_{attr}^{i}$, resulting in the signal $\tilde{g}_{attr}^{i}$. Next, we compute the mean squared error $\mathrm{MSE}(\cdot, \cdot)$ to quantify the difference between the two signals. $L_{TC-Smooth}$ can be expressed as:
\begin{equation}
L_{TC-Smooth}=\frac{1}{N_{total}} \sum_{i=1}^{N_{total}}{\mathrm{MSE}(g_{attr}^{i}, \tilde{g}_{attr}^{i})}.
\end{equation}

$\mathbf{L_{PC-CLIP}}$. $L_{PC-CLIP}$ is used to align the generated 4D content with the provided text description. Specifically, the corresponding prompt of the sequence $V$ is denoted as $tx$, and the text encoder and image encoder in the CLIP \cite{radford2021learning} model are represented as $E_{tx}$ and $E_{im}$, respectively. $L_{PC-CLIP}$ can then be expressed as:
\begin{equation}
L_{PC-CLIP}=\frac{1}{T} \sum_{t=1}^T{(1-\mathrm{cos}(E_{tx}(tx), E_{im}(V^t)))},
\end{equation}
where $\mathrm{cos}$ denotes the cosine similarity.

\textbf{Fusion Strategy.} We adopt the following formula to implement the loss function fusion:
\begin{equation}
\begin{aligned}
L=& \theta_1 L_{SC-SSIM}+\theta_2 L_{SC-Rig}+\theta_3 L_{TC-FVD}+  
\\ & \theta_4 L_{TC-Smooth}+\theta_5 L_{PC-CLIP}.
\end{aligned}
\end{equation}
Based on our experience, we set the weight parameters for the fusion as $[\theta_1, \theta_2, \theta_3, \theta_4, \theta_5]=[1, 0.01, 0.001, 0.1, 1]$.

\subsection{Implementation Details}

Here, we provide the implementation and training details of the network. During both training and inference, we randomly sample Gaussians (including their position and attributes) from standard Gaussian noise. The total number of sampled Gaussians is $N_{total}=40000$, with a total frame length of $T_A=12$. The grouping strategy clusters them into $G=400$ groups. CLIP encodes the text into a 768-dimensional vector, and the latent feature dimension during the network modeling process is set to $D’=768$. For DDIM, the total number of denoising steps is set to $M=50$, and its denoiser consists of a 6-layer Transformer.
The layer number of GroupFormer in GIE is set to $L=6$ with hidden dimension $D'=768$.
The temporal extension ratio in TED is set to $\eta=2:1$ for a better balance between performance and computational cost.
Additionally, we use 4 NVIDIA RTX 3090 GPUs for training. Adam \cite{kingma2014adam} is used as the optimizer with weight decay $5\times10^{-4}$ and an initial learning rate of $1\times10^{-4}$. 
The learning rate is reduced by a factor of $\times 0.1$ at 8K and 9K iterations, with a total of 10K training iterations. 

We further provide the matrix dimensions of each module in STP4D to aid readers in better understanding our method.

\textbf{Details in TPE.} After the process of grouping strategy, the Gaussians $\mathcal{G}' \in \mathbb{R}^{T_A \times G \times N \times D}$ are obtained. Before cross-attention modeling in TPE, this matrix is resized to $(T_A \times G) \times (N \times D)$ and mapped by MLPs to $(T_A \times G) \times D'$, ensuring that $\tilde{e} \in \mathbb{R}^{T_A \times D'}$ is effectively embedded into the Gaussians. After $M=50$ denoising steps, the Gaussian features $(T_A \times G) \times D'$ are remapped by MLPs back to $T_A \times G \times N \times D$, resulting in the Gaussians $\mathcal{G}^P$ after prompt consistency modeling.

\textbf{Details in GIE.} $\mathcal{G}^P$ is transformed into $\mathcal{G}^{P}_{1} \in \mathbb{R}^{GN \times D}$, $\mathcal{G}^{P}_{2} \in \mathbb{R}^{G T_A \times D}$, and $\mathcal{G}^{P}_{3} \in \mathbb{R}^{N T_A \times D}$ by 3D K-Planes modeling~\cite{fridovich2023k}. After processing with GroupFormer, the matrix dimensions remain unchanged. Concatenation of these matrices forms the Key and Value for the cross-attention mechanism, with a matrix size of $(GN + GT_A + NT_A) \times D$. However, directly using this matrix as Key and Value would result in an excessive token count (in the STP4D experimental setup, $GN + GT_A + NT_A$ reaches 46,000), exceeding our capacity. Therefore, we introduce a resize parameter $N^{+}$ to reshape this matrix into $\frac{(GN + GT_A + NT_A)}{N^{+}} \times (N^{+} D)$, where $N^{+}=N=100$. $\mathcal{G}^P \in \mathbb{R}^{T_A \times G \times N \times D}$ is used as the Query for cross-attention, resized to $(T_A \times G) \times (N \times D)$ and mapped to $(T_A \times G) \times D'$. After cross-attention, the geometrically enhanced Gaussians $\mathcal{G}^{SP}$ are obtained.

\textbf{Details in TED.} In the TED module, the cross-attention mechanism uses $\mathcal{G}^{SP} \in \mathbb{R}^{T_A \times G \times N \times D}$ as the Key and Value, and a learnable weight pool $P_{\mathcal{G}} \in \mathbb{R}^{T \times G \times N \times D}$ as the Query. Similar to the previous process, before cross-attention, both matrices are first resized and then mapped to the latent features with dimension $D'$. Finally, after spatio-temporal-prompt consistency modeling, the Gaussians $\mathcal{G}^{STP} \in \mathbb{R}^{T \times G \times N \times D}$ are obtained.

\subsection{Dataset Information}
Diffusion4D~\cite{liang2024diffusion4d} is a large-scale, high-quality dynamic 3D dataset collected from the Objaverse-1.0~\cite{deitke2023objaverse} and Objaverse-XL~\cite{deitke2024objaverse} public repositories. For text-to-4D tasks, Cap3D~\cite{luo2024scalable, luo2024view} is used to generate descriptive captions for each 4D asset. We utilize the Diffusion4D dataset based on Objaverse-1.0, including approximately 11,600 sets of 4D assets and their text descriptions, each comprising 24 rendered frames.


\section{Additional Experimental Results}
\label{sec:s2}

\subsection{Study of Temporal Extension Ratio}


\begin{table}[t]
    \centering
    \caption{Comparison of different temporal extension ratio $\eta$ in TED in terms of the quantitative metrics on Diffusion4D.}
    \begin{tabular}{cccc}
    \toprule
$\eta$    & CLIP-F↑ & FVD↓ & Inference Time↓ \\ \midrule
$4:1$ &  0.314       &   991.0       &    \textbf{$\sim$2.8s}             \\
$3:1$ &    0.592     &     786.5     &   $\sim$3.6s              \\
$\mathbf{2:1}$ \textbf{(Ours)} & 0.841   & 441.1    & $\sim$4.6s      \\
$4:3$ &    \textbf{0.853}     &     \textbf{432.5}     &   $\sim$6.5s  
\\ \bottomrule
\end{tabular}
    \label{tab: eta1}
\end{table}

We explore the impact of the temporal extension ratio $\eta$ on network performance within the TED module. As shown in Tab.\ref{tab: eta1}, we compare the different values of $\eta$ applied to the TED module. The results indicate that with a large $\eta$, \textit{e.g.} $\eta=3:1$ or $\eta=4:1$, the CLIP-F score drops sharply, while the FVD score rises significantly. This suggests that although the computational and time consumption for training and inference are reduced, a smaller number of anchor frames is insufficient to serve as bases, making it difficult for the actual frames to be accurately weighted from the anchor frames.
Although $\eta = 4:3$ achieves the best performance, it only slightly outperforms $\eta = 2:1$ in terms of the CLIP-F score (+0.012) and FVD score (-8.6). This minor improvement comes at the cost of a seriously higher computational burden, with inference time increasing by approximately 41\%. Therefore, we finally adopt the $\eta = 2:1$ strategy, considering the balance between time consumption and generation performance.

To further verify the generality of $\eta = 2:1$, we further generate 50-frame 4D content using different $\eta$ values on Objaverse-1.0. We select 1,000 dynamic 3D assets from the Objaverse-1.0 repository to validate the generalizability of the temporal extension ratio $\eta=2:1$ in TED. Specifically, we render a 50-frame ground truth video for each asset, while the frame length for STP4D training and inference is set to 50. We adopt different values of $\eta$ to generate 4D content, with the experimental results shown in Tab.\ref{tab: eta2}. The results indicate that $\eta=2:1$ still delivers excellent performance with a short inference time, achieving 0.625 and 745.8 in CLIP-F and FVD scores, respectively. In contrast, using a larger $\eta=10:3$ results in a significant performance drop, suggesting that the anchor frames are insufficient to serve as bases to generate actual frames in this strategy. When $\eta$ is smaller than $2:1$, such as $\eta=10:7$, the network performance slightly improves, but the inference time increases significantly by 75\%, leading to higher computational cost. In summary, setting the temporal extension ratio to $2:1$ achieves a balance between computational consumption and network performance when generating content with different frame lengths, demonstrating its generalizability.

\begin{table}[ht]
    \centering
    \caption{Comparison of different temporal extension ratio $\eta$ in TED in terms of quantitative metrics on Objaverse-1.0.}
    \begin{tabular}{cccc}
    \toprule
$\eta$    & CLIP-F↑ & FVD↓ & Inference Time↓ \\ \midrule
$10:3$ &  0.338       &   1508.5       &    \textbf{$\sim$5.8s}             \\
$\mathbf{2:1}$ \textbf{(Ours)} &    0.625     &     745.8     &   $\sim$9.6s              \\
$10:7$ &    \textbf{0.640}     &     \textbf{715.8}     &   $\sim$16.8s  
\\ \bottomrule
\end{tabular}
    \label{tab: eta2}
\end{table}

\subsection{Study of Gaussian Groups}


\begin{table}[t]
    \centering
    \caption{The impact of parameter $G$ (groups of Gaussians) in terms of the quantitative metrics on Diffusion4D.}
    \begin{tabular}{cccc}
    \toprule
$G$    & CLIP-F↑ & CLIP-O↑ & FVD↓ \\ \midrule
200 &   0.828       &  0.695   &  447.8             \\
\textbf{400} \textbf{(Ours)}&    \textbf{0.841}     &     \textbf{0.705}    &   \textbf{441.1}              \\
500  & 0.835   & 0.698    & 443.6     \\
800 &    0.815    &   0.673   & 460.5  
\\ \bottomrule
\end{tabular}
    \label{tab: G}
\end{table}

The grouping strategy is crucial for the modeling process of Gaussians and enforcing local rigidity constraints. Therefore, we perform a parameter analysis on the number of groups $G$, as shown in Tab.\ref{tab: G}. The results indicate that, given the same total number of Gaussians, dividing them into $G=400$ groups outperforms other grouping strategies across all metrics. As the number of groups increases or decreases, network performance declines. This is because with too many groups, the number of Gaussians per group decreases, leading to less information within each group, which negatively affects the geometric modeling process in GIE. Conversely, with too few groups, the local rigidity becomes too loose and may fail to provide sufficient constraints, damaging spatial consistency. Therefore, both overly large and small group sizes compromise local rigidity constraints and spatial consistency. Additionally, an unreasonable grouping strategy may consume excessive resources during the modeling process, preventing the model from training effectively.

\begin{figure*}[htbp]
    \centering
    \includegraphics[width=0.95\linewidth]{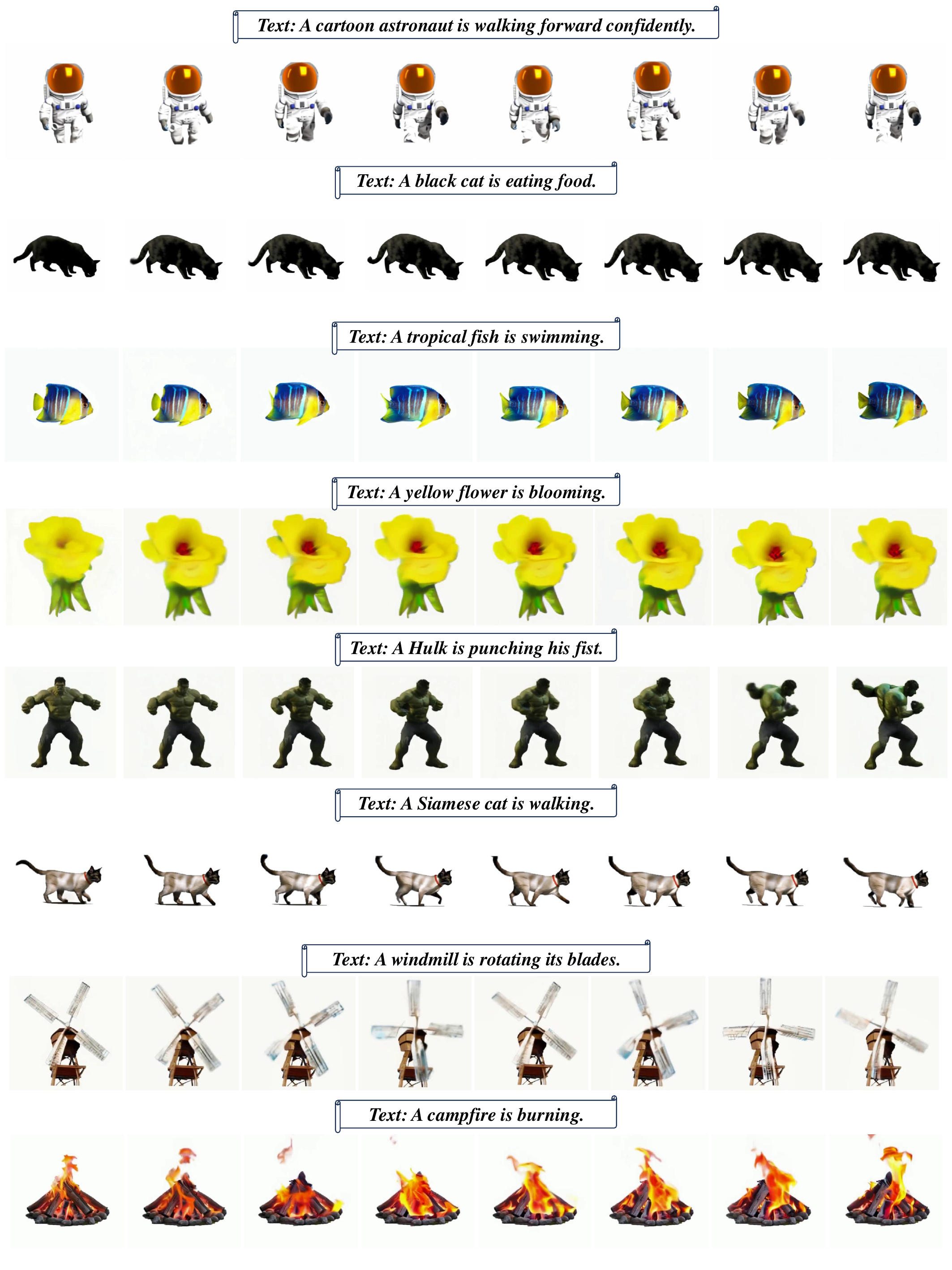}
    \caption{Various visualizations of 4D assets generated from STP4D.
    }
    \label{fig: appendix}
\end{figure*}

\subsection{Additional Visualizations}

Furthermore, we present additional visualizations of 4D assets generated by STP4D, as shown in Fig.\ref{fig: appendix}. 
Additionally, \textbf{GIF files} corresponding to all 4D assets in the paper are available in the \textbf{attachments of the submitted supplementary materials}.

\section{Discussion}
\label{sec:s3}

\subsection{Progressive Design of STP4D}

STP4D adopts a progressive design approach, where each module is driven by a clear design motivation. At the beginning, we used a simple text embedding method and applied DDIM to denoise the Gaussian attributes for generating 4D content. We then found that embedding different text information frame by frame into the Gaussians improved content-text alignment, as more targeted text features for each frame lead to more accurate and expected results. Based on this, we designed the TPE module to ensure a high alignment between generated content and text descriptions. After achieving stable 4D content generation with TPE and DDIM, we focused on enhancing spatial consistency. Given that Gaussians are modeled in groups, we introduced GroupFormer to model the spatio-temporal relationships within and between Gaussian groups, and proposed a local rigidity loss to enforce spatial consistency. To achieve efficient spatio-temporal modeling, we further incorporated the K-Planes technique before GroupFormer to decouple the high-dimensional Gaussians into low-rank matrices, which are suitable for modeling. However, due to GPU resource limitations, our method initially supported only very short 4D generation. To address this, we extended the temporal frames by using existing frames as anchors, applying loss functions to ensure temporal consistency and generation quality. In summary, this progressive design approach allows us to address existing challenges from multiple dimensions, guiding the network design with clear motivations and enhancing performance to achieve satisfactory results.

\subsection{Potential in Complex Scenarios}

While our method achieves impressive performance and speed in some simple scenes, generating assets in complex scenes still remains challenging. This limitation mainly arises from the constraints of training data. Specifically, our method was trained exclusively on Diffusion4D, the only open-source text-to-4D dataset available at the time, which contains 11,600 simple 4D assets with corresponding texts. Our method faces challenges when generating assets in complex scenes due to the lack of complex 4D training data. Therefore, we believe that with richer datasets and a larger network, our method could achieve better results in complex scenarios.

On the other hand, limited GPU resources restricted the training of STP4D to only 40,000 Gaussians. With this relatively small number of Gaussians, the model’s ability to represent complex scenes was constrained. Forcing a fixed number of Gaussians to fit complex scenes (such as those requiring high resolution) leads to blurriness and loss of detail, as Gaussians must cover larger areas, rather than capturing finer details. Therefore, the limited number of Gaussians is another factor which causes STP4D's suboptimal performance in complex scenarios.

\bibliographystyle{stp4d}
\bibliography{stp4d}

\end{document}